# "Genlangs" and Zipf's Law:
# Do languages generated by ChatGPT statistically look human?

Justin Diamond and ChatGPT
29 March 2023

**Abstract.** OpenAI's GPT-4 is a Large Language Model (LLM) that can generate coherent constructed languages, or "conlangs," which we propose be called "genlangs" when generated by Artificial Intelligence (AI).[1] The genlangs created by ChatGPT for this research (Voxphera, Vivenzia, and Lumivoxa) each have unique features, appear facially coherent, and plausibly "translate" into English.[2] This study investigates whether genlangs created by ChatGPT follow Zipf's law.

Zipf's law approximately holds across all natural and artificially constructed human languages.[3,4] According to Zipf's law, the word frequencies in a text corpus are inversely proportional to their rank in the frequency table. This means that the most frequent word appears about twice as often as the second most frequent word, three times as often as the third most frequent word, and so on.

We hypothesize that Zipf's law will hold for genlangs because (1) genlangs created by ChatGPT fundamentally operate in the same way as human language with respect to the semantic usefulness of certain tokens, and (2) ChatGPT has been trained on a corpora of text that includes many different languages, all of which exhibit Zipf's law to varying degrees.

Through statistical linguistics, we aim to understand if LLM-based languages statistically look human. Our findings indicate that genlangs adhere closely to Zipf's law, supporting the hypothesis that genlangs created by ChatGPT exhibit similar statistical properties to natural and artificial human languages.

We also conclude that with human assistance, AI is already capable of creating the world's first fully-functional genlang, and we call for its development.

**Acknowledgements.** I extend thanks to Jonny Miller and Jean-Paul R. Soucy for raising the question addressed in this paper.[5] Although the substance of this paper was written entirely by a human, ChatGPT contributed by generating innovative languages with a large corpora of text and assisting in the creation of matplotlib figures cited throughout. Accordingly, "we" is the pronoun used throughout this paper. Any text written by ChatGPT will be highlighted in green.

**Introduction.** On March 14th, 2023, OpenAI released GPT-4, which is presumably the largest and most advanced LLM to date.[6] With hundreds of billions of parameters, GPT-4 is capable of performing a wide range of tasks, including but not limited to language translation, question-answering, and computer programming.

---

ChatGPT is now built on GPT-4, and its general applicability to intelligence-intensive tasks and incredible usefulness has garnered much attention.

Intriguingly, ChatGPT can also be prompted to generate coherent and convincing artificial languages which in reality do not exist. These LLM-based languages reflect constructed languages, or "conlangs," which have been systematically invented by individuals or groups rather than having developed naturally over time. Examples of conlangs include Esperanto, which was invented by L. L. Zamenhof in the late 19th century, and Sindarin and Quenya which were invented by J.R.R. Tolkien for the Lord of the Rings trilogy in the mid 20th century.[7]

Creating a comprehensive conlang has historically required vast amounts of human effort, dedication, and time. However, new LLMs like ChatGPT have the novel ability to generate facially coherent conlangs with imaginative rules and consistent functionality in seconds. Simple and open-ended prompts can now be used to coax LLMs into generating original "languages." Given the distinction between earlier conlangs and recent innovations in generated artificial language, we propose that AI generated languages be uniquely referred to as "Genlangs" for the purposes of future research and discussion.

An obvious starting point for the study of LLM-based genlangs is to: (1) Use LLMs to create novel genlangs, and (2) Probe these languages with a series of tests regarding their syntactic, phonological, morphological, semantic, pragmatic, and other such properties. In this study, we focus only on whether genlangs created by ChatGPT yield a corpora of text that holds to "Zipf's law," like natural and artificial human languages.

Zipf's law is an empirical law discovered by George Kingsley Zipf in 1936.[8] Zipf's law states that in a corpus of text, the frequency of any word is inversely proportional to its rank in the frequency table. In other words, the most frequent word appears about twice as often as the second most frequent word, three times as often as the third most frequent word, and so on. Expressed mathematically, the $r$th most frequent word has a frequency $f(r)$ that scales according to

$$f(r) \propto \frac{1}{r^\alpha}$$

for $\alpha \approx 1$.[9] This implies that across languages, the distribution of words is expected to approximately follow a mathematical power law.

Zipf's law identifies the fact that across human languages, high-frequency words which account for most of the tokens in text (e.g., "a," "the," "I," etc.) are used in roughly predictable proportions to low-frequency words (e.g., "accordionist," "newfangled," "toucan," etc.).

This relationship between words has been observed to approximately hold in most corpora of text across all natural and artificially constructed human languages.[10] For this reason, Zipf's law is often used in the study of statistical linguistics. Although the law is an empirical observation, and not a universal law that will hold true in all circumstances, it is still useful in a handful of ways. With respect to LLMs, Zipf's law can be used to identify whether genlangs *statistically* look like human languages.

**Question.** Do languages generated by ChatGPT statistically look human? Whether genlangs will hold to Zipf's law remains largely, if not entirely, unexplored. Answering this question will shed light on the extent to which genlangs conform to the statistical regularities of natural and artificial human language. The ultimate purpose of this inquiry is to improve our understanding of technology and linguistics, while informing the development of LLMs and LLM-based genlangs.

---

[7] Campisi, Estephan G. "Babel from Scratch: Examining the Impact and Usage of Constructed Languages in Literature ." PDXScholar . Portland State University, March 1, 2018. https://pdxscholar.library.pdx.edu/.

[8] Zipf, George Kingsley. The psycho-biology of language: An introduction to dynamic philology. Vol. 21. Psychology Press, 1999.
[9] Piantadosi, Steven T. "Zipf's Word Frequency Law in Natural Language: A Critical Review and Future Directions." Psychonomic Bulletin & Review. U.S. National Library of Medicine, October 2014. https://www.ncbi.nlm.nih.gov/pmc/articles/PMC4176592/.
[10] Ibid.

**Hypothesis.** We hypothesize that the statistical properties of natural and artificial human languages will be reflected in genlangs newly created by ChatGPT. This is expected because (1) genlangs created by ChatGPT will fundamentally operate in the same way as human language with respect to the semantic usefulness of certain tokens, and (2) ChatGPT has been trained on a corpora of text that includes many different languages, all of which exhibit Zipf's law to varying degrees.

**Methodology.** In order to test whether LLM-based genlangs follow Zipf's law, we will start by giving ChatGPT (Model GPT-4) the following prompts:

*Prompt 1:* You are "Lang-GPT" – your ambitious task is to invent a brand new language with innovative rules. First, what is the name of this language? Second, write a few paragraphs in this new language with a translation below. Third, describe how this language works in terms of grammar, syntax, and any other relevant rules.

*Prompt 2:* Can you write me a short story in [Genlang] and provide a translation?

*Prompt 3:* Can you write a story that is as long as possible in [Genlang]?

The second and/or third prompt will be repeated until a corpus of text at least 1,000 words long is yielded for each genlang. A corpus of this size should be enough to observe whether a Zipfian or near-Zipfian distribution appears. Three genlangs and their respective English translations will be created and analyzed in this study. All text corpora produced will be viewable at the link cited here.[11]

Next, the texts of each genlang will be combined into a single corpus of text. This large corpus of consolidated genlangs will reflect the "sum" of their unique properties, and is meant to be compared with ChatGPT generated English. ChatGPT will be asked to generate English via a series of prompts similar to those above (e.g. "Can you write me a short story in English?" and "Can you write as much text as possible in English?").

Finally, we will collect text from (1) a human-authored English novel, (2) ChatGPT generated "gibberish," and (3) randomly generated English words. The human-authored English novel that will be tested is the Project Gutenberg version of "Tom Swift And His Electric Runabout," by Victor Appleton.[12] The ChatGPT generated gibberish will be created with the prompt: "Can you generate some random gibberish for me? Write as much random gibberish as possible." Lastly, the randomly generated English words will be produced with an online tool that outputs any number of random words, with duplicates.

To avoid bias in the process of text generation, no responses or words will be filtered out except unrelated hallucinations or niceties of the chatbot. The first three genlangs created will be the only ones studied. Once all texts have been generated, collected, and/or compiled, their Zipfian distributions and correlations will be calculated and visualized for comparison.

**Linguistic Findings.** When ChatGPT was prompted to create genlangs, write texts in those genlangs, and provide translations/explanations for these texts, it invented the following languages…

---

[11] Justin Diamond. "Justin-Diamond/Genlangs-and-Zipfs." GitHub. March 29, 2023.
https://github.com/Justin-Diamond/genlangs-and-zipfs.
[12] Appleton, Victor. "Tom Swift and His Electric Runabout." Project Gutenberg, July 13, 2008.
https://www.gutenberg.org/files/950/950-h/950-h.htm.

## Voxphera:

*Response to Prompt 1:*
The name of this language is "Voxphera."

In Voxphera:
Bintaz orensta kishalim, glavnis meyol tivanto. Sankhara zelindor freyata hikvona, chofta kenla nivarep. Trinkalos golvinda, ulfaro pinthos belnari, drimka vespilon ankares.

Translation:
In the lush forest, birds sang harmoniously. The wind whispered gently through the leaves, carrying a scent of renewal. Amidst the serene beauty, a sense of belonging enveloped the soul.

Language rules:
Grammar: Voxphera is an inflecting language, where the function of a word in a sentence is determined by its ending. Nouns, adjectives, and verbs are all inflected for case, gender, and number. There are four cases (nominative, genitive, dative, and accusative), three genders (masculine, feminine, and neuter), and two numbers (singular and plural).
Syntax: The basic word order of Voxphera is subject-verb-object (SVO). However, due to the inflections, the word order can be quite flexible. Adjectives agree with the nouns they modify in case, gender, and number and typically precede the noun. Adverbs typically follow the verb but can be placed elsewhere for emphasis.
Verb system: Verbs in Voxphera are conjugated for tense, aspect, and mood. There are three tenses (past, present, and future), two aspects (perfect and imperfect), and four moods (indicative, subjunctive, imperative, and optative). Verbs also agree with their subjects in person and number.
Phonology: Voxphera has a rich phonemic inventory, including a variety of consonants, vowels, and diphthongs. The language has a melodic quality, with stress typically falling on the penultimate syllable.
Writing system: Voxphera uses an alphabetic script consisting of 24 letters. It is written from left to right, with spaces between words and punctuation to indicate sentence boundaries.
Vocabulary: The vocabulary of Voxphera is derived from a combination of natural sources and creative imagination, aiming to evoke a sense of wonder and connection with the world around us.

*Response to Prompt 2:*
Krenos, flornik yalvondri, tovasta melinora yonkalim preniva. Zuntra zovetis, mivaro entalos dimbri fenkaro. Tarnok grinfal, onkresa yamora tevindo, draksta kizula elnari. Oflanis jeritri, shabris dranis, krenos ilvanki arbloros. Silvanis, vellari hikvona, marenta orinika krenos. Bintaza frovinda, silvanis krenos thalara eplosta, wintara yalvondri mekora.

Translation:
Krenos, a young adventurer, wandered through the enchanting forest of Preniva. The sun filtered down, casting a golden light upon the verdant landscape. Along the way, he encountered a magnificent creature – a shimmering dragon. With a heart full of courage, Krenos approached the majestic being. Silvanis, guardian of the forest, gazed back at Krenos. With a gentle gust of wind, Silvanis lifted Krenos onto its back, soaring through the vibrant skies together.

## Vivenzia:

*Response to Prompt 1:*
The name of this new language is "Vivenzia."

Here are a few paragraphs in Vivenzia, followed by their translations:

Vivenzia:
"Zovento kila prazuflo, ta farni uzaq hevela marq. Yora nivrali ezumi, eno zirka yaldra pelto. Sodwe lixta vureja, miq vana trelesta delmor."

Translation:
"In the beautiful forest, a small bird sang a joyful tune. The sun's rays illuminated the leaves,

creating a magical atmosphere. Nearby, a gentle stream flowed, carrying the sweet scent of flowers."

Now let's explore the language rules of Vivenzia.

Grammar and Syntax:
Word order: Vivenzia follows a Subject-Verb-Object (SVO) word order, similar to English.
Adjectives: Adjectives are placed before the noun they modify.
Verbs: Verbs do not conjugate based on tense or subject agreement. Tense is expressed through time markers placed at the beginning of a sentence.
Nouns: Nouns do not have grammatical gender or plural forms. The context and number words convey the quantity.
Prepositions: Prepositions are used to express relationships between words, similar to English.
Pronouns: Personal pronouns are used, and they decline for subject, object, and possessive forms.
Additional Rules:
Syllable structure: Vivenzia favors consonant-vowel (CV) syllable structure. This makes it easy to pronounce and learn.
Base-20 numbering system: Vivenzia uses a base-20 numbering system, unlike the base-10 system used in many other languages.
Time markers: To express tense, Vivenzia uses time markers at the beginning of a sentence. "Zovento" (past), "Prenso" (present), and "Futuro" (future) are the three primary time markers.
Logical connectors: Vivenzia uses logical connectors to combine ideas and sentences, such as "eno" (and), "kliri" (but), and "orev" (or).

In summary, Vivenzia is an innovative language that employs a simple structure, making it easy to learn and pronounce. Its unique features, like the base-20 numbering system and time markers, set it apart from other languages.

*Response to Prompt 2:*
Vivenzia:
"Prenso kavira lumoza, zimla xarela qinbran ala nortu. Jiklo wivu, vovra yaltra, eno gizlan kruno aya bistra. Pivu jiklo yentro dulara ziblon. Yora strina jiklo eno yaltra kanto zovro. Eno zirka, sodwe lixta skavu eno vurna drila.

Zovento jiklo bivran kruno, ta vovra yaltra hizla qalro. Kliri jiklo yaltra ala yora nivrali, eno zirka sodwe lixta. Jiklo trexlo, eno yaltra qula. Orekta yentro dulara, vovra yaltra ziblon. Kliri yora strina jiklo eno yaltra, eno marq skavu eno vurna drila.

Futuro jiklo yaltra, eno gizlan kruno, zivlo zenu vurna drila. Vovra yaltra yora strina kanto vena, eno marq kruno eno skavu. Eno yora nivrali, jiklo eno yaltra zirka zovro eno yora strina."

Translation:
"Once upon a time, in a distant land far to the north, there lived a young boy named Jiklo, his sister Yaltra, and their wise grandfather in a small village. One day, Jiklo ventured into the enchanted forest. The sun shone through the trees and the birds sang all around. Nearby, a gentle stream flowed, and colorful flowers bloomed.

Long ago, Jiklo's grandfather had told them stories of the magical forest, where the sun's rays could reveal hidden treasures. Despite the warnings, Jiklo and Yaltra ventured deeper into the sunlit woods. They laughed, played, and eventually stumbled upon a mysterious cave. Heeding their grandfather's tales, they carefully entered the cave. Inside, they found a magical gemstone that had the power to heal and protect the village.

In the future, Jiklo, Yaltra, and their grandfather used the gemstone's powers to keep their village safe and prosperous. The sun continued to shine on the land, filling it with life and magic. In the heart of the enchanted forest, Jiklo and Yaltra would often return to play and explore, sharing their adventures with the village."

## *Lumivoxa:*

*Response to Prompt 1:*
The name of this new language is "Lumivoxa."

Here is a sample paragraph in Lumivoxa:

Lumivoxa:
"Zemira nar klovar u-tayra, grika frova zintis plamark. Jikil krevos blanu o-manti, shterpa prindel wadli yonex. Thavro qarik mepro gweftan, bimlo shtark dufro vori."

Translation:
"In the land of beautiful mountains, the trees danced with the wind. The river flowed calmly through the valley, nourishing the creatures that dwelled nearby. The sun illuminated the landscape, casting warm light on every corner."

Grammar, Syntax, and Relevant Rules:
Word order: Subject-Verb-Object (SVO), similar to English.
Noun gender: Lumivoxa has three genders: feminine, masculine, and neuter. Gender is indicated by the first letter of the noun: "Z" for feminine, "J" for masculine, and "T" for neuter.
Noun cases: Nominative (subject), accusative (object), and genitive (possession). The nominative case is unmarked, the accusative case adds the prefix "u-" to the noun, and the genitive case adds the prefix "o-".
Adjectives: Adjectives follow the noun they modify and agree with the noun's gender by taking the same initial letter as the noun.
Verbs: Verbs are not conjugated for person or number. Instead, tense is indicated by the final letter of the verb: "a" for past, "i" for present, and "u" for future. Additionally, the verbs are prefixed with "p" for the first person, "g" for the second person, and "s" for the third person.
Plurals: Plurals are formed by adding the suffix "-s" to the noun, regardless of gender.
Prepositions: Prepositions are used to indicate relationships between words, similar to English. They precede the noun they relate to and take the form of a single syllable.

This innovative language, Lumivoxa, combines familiar elements with unique rules to create a rich and expressive means of communication.

*Response to Prompt 2:*
Lumivoxa:
"Jikil zorba nar vineti, o-sorin krevos zafyra blanu. Gikri zilana topris gweftan, yalmi jikil krevos ziblara marnu. Tadriki jikil plamark vori, melka zilana zonari. Marnu kroun, grika frova nar klovar jikil yalora. U-droni topris nar zilana zonari, wadli yonex o-manti shterpa prindel. Traklo yalmi jikil krevos ziblara, dufro vori nar mepro qarik. Jikil krevos zafyra traklo, o-sorin kroun gweftan."

Translation:
"Once upon a time in a small village, a young boy named Zafyra lived. He loved to explore the lush forests, often befriending the wild animals he encountered. One day, while chasing a colorful butterfly, he stumbled upon a hidden cave. Inside, the trees gave way to a secret world. A crystal-clear lake reflected the light of the glowing plants that surrounded it. Curiosity sparked within Zafyra, and he ventured deeper into this magical place. As he walked, he discovered a source of unparalleled beauty, forever changing his life."

This short story in Lumivoxa demonstrates the language's capacity for conveying a narrative in a unique and expressive manner.

Voxphera, Vivenzia, and Lumivoxa each have unique features, appear facially coherent, and plausibly "translate" into English. Although ChatGPT made some mistakes in applying the rules of its own genlangs, the stories and texts provided for all three *tend* to be stylistically similar and structurally consistent.

With respect to linguistics, what are some noteworthy differences and commonalities between Voxphera, Vivenzia, and Lumivoxa?

Rules like verb conjugation apply differently across all three genlangs. For example, in Voxphera, verbs are conjugated for tense, aspect, and mood; in Vivenzia, verbs do not conjugate and tense is expressed through time markers at the beginning of the sentence; in Lumivoxa, verbs are not conjugated, but tense is indicated by the final letter of the verb instead of a time marker.

Other novel quirks include Voxphera's 24-letter alphabet, Vivenzia's base-20 numbering system, and Lumivoxa's use of prefixes for noun-cases – though these features do not necessarily manifest themselves in the text corpora.

Although these genlangs are unique phonologically, morphologically, alphabetically, etc., they also share some commonalities with each other. For example, all three genlangs follow a Subject-Verb-Object (SVO) syntax like English. This may indicate that the rules "preferred" by an LLM are naturally rooted in the languages it is trained on, and when prompted to create a new genlang, an LLM could be more probabilistically likely to change word-rules (e.g., prefixes or pronouns) than sentence-rules (e.g., syntax).

**Zipfian Analysis.** After prompting ChatGPT to provide a larger corpus of text in each genlang, Zipfian distributions were derived. Zipf's law plots representing ChatGPT generated English, a human-authored English novel, ChatGPT generated gibberish, and random English words were also created for comparison and reference.

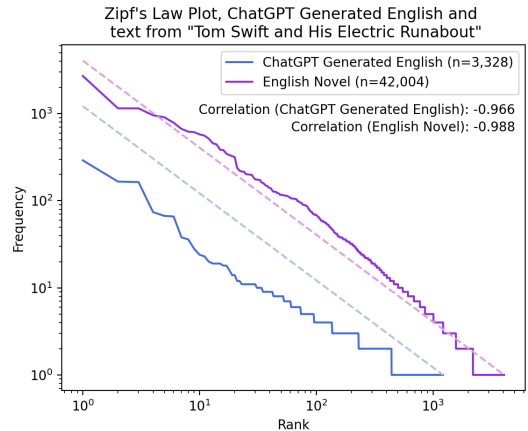

**Fig. 1.** Zipf's Law Plot, ChatGPT Generated English and an English Novel

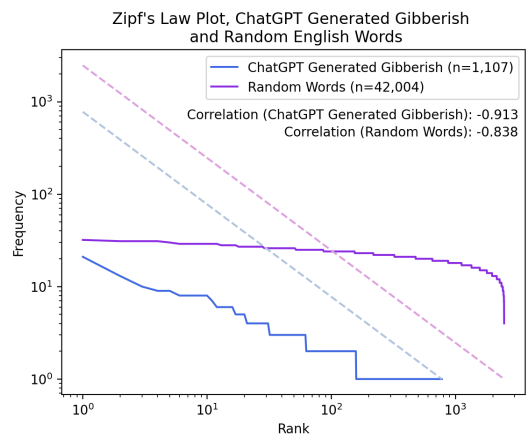

**Fig. 2.** Zipf's Law Plot, ChatGPT Generated Gibberish and Random English Words

ChatGPT generated English and the text from "Tom Swift and His Electric Runabout" both have high correlations with Zipf's law (-0.966, -0.988). The larger *n* of words in "Tom Swift and His Electric Runabout" likely accounts for its higher Zipfian correlation.

ChatGPT generated "gibberish" had some logic to its structure (e.g., four occurrences of "blibber-blabber"), but still retained a fairly low Zipfian correlation (-0.913). In contrast, the text file populated with completely random words entirely failed to follow Zipf's law, and had the lowest Zipfian correlation of all (-0.838).

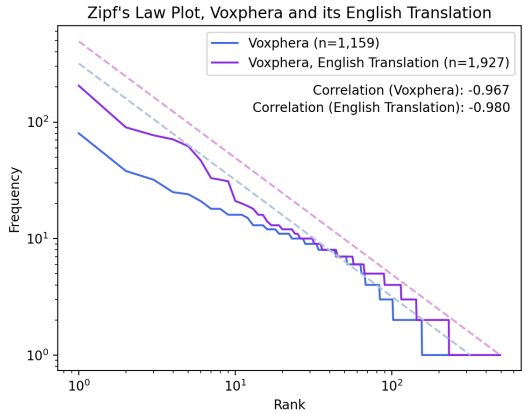

**Fig. 3.** Zipf's Law Plot, Voxphera and its English Translation

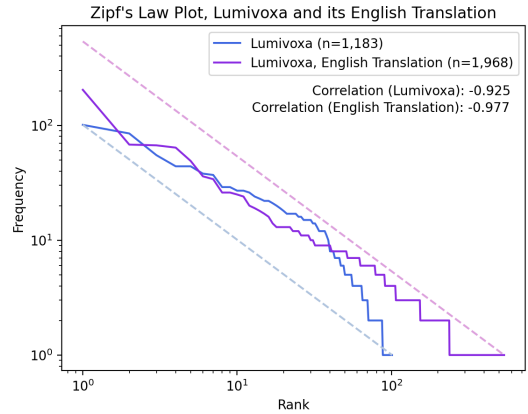

**Fig. 5.** Zipf's Law Plot, Lumivoxa and its English Translation

Of the genlangs, Voxphera and its English translation have the highest correlation with Zipf's law (-0.967, -0.980). Voxphera and ChatGPT generated English have nearly identical Zipfian correlations (-0.967, -0.966).

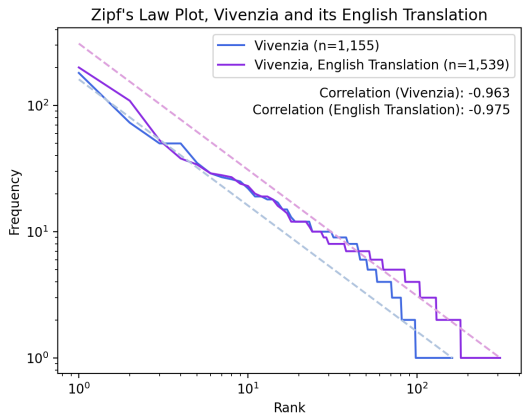

**Fig. 4.** Zipf's Law Plot, Vivenzia and its English Translation

Vivenzia and its English translation also have high correlations with Zipf's law (-0.963, -0.975). Vivenzia's correlations with Zipf's law (-0.963) are similar to Voxphera's (-0.967), thus Vivenzia also has a nearly identical Zipfian correlation to ChatGPT generated English (-0.966).

Among genlangs, Lumivoxa is the statistical outlier. It has the weakest correlation with Zipf's law (-0.925) of any coherent language observed in this paper. This correlation is closer to ChatGPT generated gibberish (-0.913) than to other genlangs or English text corpora. However, Lumivoxa behaves differently from gibberish in that its low correlation is caused by the high-frequency use of a small number of middle-rank words (causing its Zipfian curve to bend outwards), while the low correlation of gibberish is caused by the infrequent use of a large number of words (causing its Zipfian curve to bend inwards).

Lumivoxa does not necessarily "break" Zipf's law; in fact, its highest and lowest frequency words follow the law almost exactly. However, the middle-rank words were used frequently enough to cause this Zipfian curve to deviate far from what is expected. Interestingly, the Zipfian curve for Lumivoxa's English translation looks similar to those of genlangs' translations, and its correlation is roughly the same (-0.977).

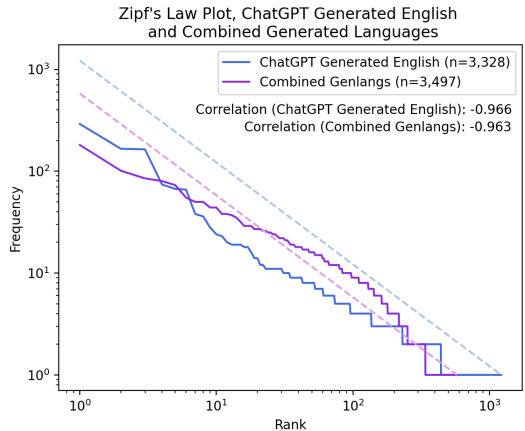

**Fig. 6.** Zipf's Law Plot, ChatGPT Generated English and Combined Genlangs

Genlangs and ChatGPT generated English have nearly identical correlations to Zipf's law (-0.966, -0.963). This indicates that genlangs and generated English follow the word frequency distribution predicted by Zipf's law to a similar degree.

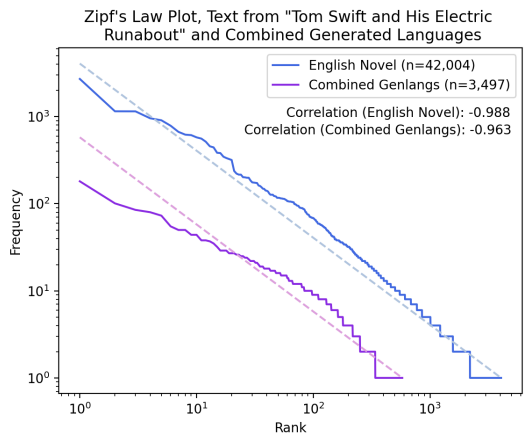

**Fig. 7.** Zipf's Law Plot, An English Novel and Combined Genlangs

Figure 7 shows that the "shape" of the combined genlangs' Zipfian curve looks strikingly similar to that of a human-authored English novel. In fact, the curve derived from genlangs looks more similar to the English novel than any other body of text – including ChatGPT generated English.

The combined genlangs and English novel also have similar Zipfian correlations (-0.963, -0.988), and the large *n* of words in "Tom Swift and His Electric Runabout" likely accounts for any difference between these correlations.

**Discussion.** Figures 6 and 7 allow us to answer the question "Do genlangs follow Zipf's law?" with a resounding "YES." Like English, genlangs created by GPT-4 follow Zipf's law. Therefore, in at least one way, genlangs *statistically* look like natural and artificial human languages.

Two thirds of observed genlangs, and genlangs in their totality, held about as closely to Zipf's law as the English language. Voxphera and Vivenzia reflected strong correlations (≥-0.963) with the ideal Zipfian distribution. However, Lumivoxa showcased the possibility of "outlier" genlangs, which only somewhat follow Zipf's law. Having a weaker correlation (-0.925), Lumivoxa showed that genlangs can also sit at the outer-bounds of language in terms of Zipfian distributions. In terms of Zipf's law, this places Lumivoxa somewhere between human languages and ChatGPT generated gibberish/completely random words.

We hypothesized that the statistical properties of natural and artificial human languages would be reflected in genlangs because (1) genlangs created by ChatGPT fundamentally operate in the same way as human language with respect to the semantic usefulness of certain tokens, and (2) ChatGPT has been trained on a corpora of text that includes many different languages, all of which exhibit Zipf's law to varying degrees.

Although this study was sufficient to show that Zipf's law holds for genlangs, our findings are not sufficient to show whether this law holds due to its linguistic "fundamentalness," or if it holds due to the fact that ChatGPT has been trained on human languages for which Zipf's law holds.

Genlangs' adherence to Zipf's law emphasizes the potential of AI technologies such as ChatGPT to simulate linguistic phenomena. As LLM research advances, we encourage a deeper exploration of the linguistic properties of genlangs. For example, future research may focus on the extent to which genlangs' Zipf curves can be altered by modifying the LLM's underlying training data.

Given the coherence of genlangs currently being created by ChatGPT, we believe that with human assistance, AI technology is already capable of developing the world's first fully-functional genlang. It is beyond the scope of this paper to speculate about which attributes this genlang should have; nonetheless we believe that genlangs could have many applications and uses if optimized in the right ways.

**Conclusion.** Our results show that genlangs created by ChatGPT adhere to Zipf's law, and the word frequency patterns observed in genlangs closely resemble those of natural and artificial human languages. These results also demonstrate the capability of modern LLMs to generate coherent, innovative languages with consistent functionality.

Conlangs of the 19th and 20th century took years of human effort, dedication, and time to develop. Like Esperanto, some were popular and well known, but none ever became widely adopted. Perhaps it was a mistake to believe that *artificial* languages could be developed by *human* intelligence… But the world is changing quickly, and the landscape of language looks different than ever before. For the first time in history, we have the opportunity to discover what happens when *artificial* languages are developed by *artificial* intelligence.

New technology can be scary, and few are comfortable venturing into the unknown. We know little about what the first fully-functional genlang will look like, and few will ever learn to speak it. But we do know this: it will follow Zipf's law.

*Esperanto:* "Lingvo internacia de la venontaj generacioj estos sole kaj nepre nur lingvo arta."[13]

*Voxphera*: "Yarenis skirona olgentar naxendara saletara trintin vinalo, artolis prenara solenta venostara."

*Vivenzia:* "Yura internazali kasta generatsi futura, jikto eno neseso, vara lingua kalarti."

*Lumivoxa:* "Jikil prestri yorvo terilis yonex, zikri plendri nari vaxlo shtormi yorvo jikil kreshti."

*English:* "The international language of the next generations will be solely and necessarily a language of art." – L.L. Zamenhof

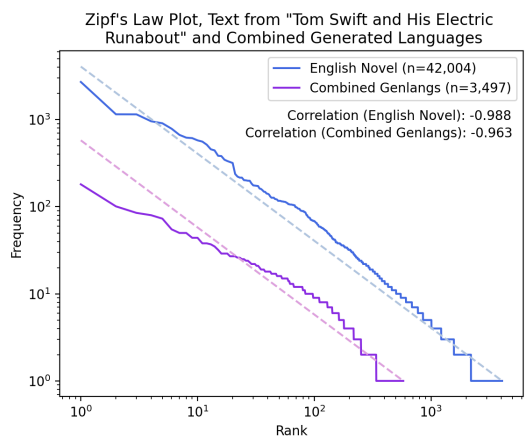

**Fig. 7.** Zipf's Law Plot, An English Novel and Combined Genlangs

**Appendix.** Cited here is a public repository which contains all data and source code related to this paper.[14] All prompts, text corpora, and code can be found in the folder labeled "ZipfGPT."

---